\documentclass{article}

\usepackage[utf8]{inputenc}
\usepackage[T1]{fontenc}
\usepackage{hyperref}
\usepackage{url}
\usepackage{booktabs}
\usepackage{amsfonts}
\usepackage{amsmath,amssymb}
\usepackage{nicefrac}
\usepackage{microtype}
\usepackage{xcolor}
\usepackage{graphicx}
\usepackage{algorithm}
\usepackage{algorithmic}
\usepackage{enumitem}
\usepackage{multirow}
\usepackage[preprint]{neurips_2026}

\newcommand{\tablestyle}[2]{\setlength{\tabcolsep}{#1}\renewcommand{\arraystretch}{#2}\centering\footnotesize}

\title{TTF: Temporal Token Fusion for \\ Efficient Video-Language Model}

\author{%
  Simin Huo \\
  Shanghai Jiao Tong University \\
  \And
  Ning Li \\
  Shanghai Jiao Tong University \\
}

\begin{document}

\maketitle

\begin{abstract}
  Video-language models (VLMs) face rapid inference costs as visual token counts scale with video length. For example, 32 frames at $448{\times}448$ resolution already yield >8,000 visual tokens in Qwen3-VL, making LLM prefill the dominant throughput bottleneck. Existing methods often rely on global similarity or attention‑guided compression, incurring offsets to their gains. We propose \textbf{Temporal Token Fusion (TTF)}, a training-free, plug-and-play pre-LLM token compression framework that exploits structured temporal redundancy in video. TTF automatically selects an anchor frame, then for each subsequent frame, performs a local window similarity search (e.g.,$3\times 3$), fusing tokens that exceed a threshold. The compressed sequence maintains positional consistency across both prefill and decoding through coordinate realignment, enabling seamless integration with existing VLM pipelines. On Qwen3-VL-8B with threshold t=0.70, TTF removes about 67\% of visual tokens while retaining 99.5\% of the baseline accuracy and introducing only ${\approx}0.16$\,GFLOPs of matching overhead. Overall, TTF offers a practical, efficient solution for video understanding. The code is available at \href{https://github.com/Cominder/ttf}{https://github.com/Cominder/ttf}
\end{abstract}

\section{Introduction}
\label{sec:intro}

Video-language models (VLMs) have become the workhorse for long-form video
understanding~\cite{videomme,mlvu,mvbench}.
They concatenate frame-wise visual tokens from a vision encoder with text and feed the
joint sequence to a decoder-only language model.
Because each frame contributes hundreds of tokens, even moderate frame budgets produce
thousands of visual tokens before any text is appended.
Under Qwen3-VL~\cite{qwen3vl}, 32 frames at $448{\times}448$ with a $2{\times}2$ spatial merge
yield \textbf{8{,}192} visual tokens per clip, so LLM prefill and key--value cache growth
dominate latency and limit throughput in interactive settings.

Training-free \emph{token compression} is a natural response.
Methods differ in \emph{where} compression is applied.
\emph{Intra-LLM} pruning~\cite{fastv,sparsevlm,pdrop} uses attention inside early transformer
layers to drop tokens, but requires materialising attention weights and is therefore often
incompatible with memory-efficient attention kernels~\cite{flashattn2}; peak memory can even
rise relative to the uncompressed model~\cite{vidcom2}.
\emph{Pre-LLM} methods~\cite{fastervlm,visionzip,dycoke,vidcom2} shorten the sequence before
the language model and preserve kernel compatibility.
Yet existing pre-LLM designs face a three-way challenge.
Frame-independent selectors~\cite{fastervlm,visionzip} treat each frame as an isolated image
and discard cheap cross-frame coherence.
Temporal clustering approaches~\cite{dycoke,holitom} group tokens across windows but apply
uniform intensity without spatial correspondence.
Matrix-based token merging~\cite{tome,mame}---which achieves strong results on single images
by computing a global $N/2{\times}N/2$ pairwise similarity matrix---scales poorly to video:
matching each of $(F{-}1)$ source frames against an anchor requires $(F{-}1)N^2$ dot products,
exceeding $2{\times}10^6$ operations for $F{=}32$, $N{=}256$ \emph{before any LLM computation},
and the cost grows quadratically with resolution.

We observe that natural video exhibits \emph{pointwise temporal redundancy}: the same scene
region tends to reappear at nearly the same grid location across nearby frames when motion is
small relative to the token grid.
TTF exploits this by comparing each source frame token only to a small spatial neighbourhood $r\times r$ in a
chosen \emph{anchor} frame, selecting the best match by cosine similarity, and fusing tokens
that exceed a single threshold~$t$.
The cost is $\mathcal{O}(9FNC)$ with default $3{\times}3$ search---linear in the number of
tokens for this pass (vision encoding and LLM attention remain the dominant terms end-to-end), and independent of any full pairwise similarity matrix.

\paragraph{Contributions.}
\begin{itemize}[leftmargin=1.4em, topsep=2pt, itemsep=1pt]
  \item \textbf{Temporal Token Fusion}: a training-free and plug-and-play pre-LLM token compression method with $\mathcal{O}((2r{+}1)^2 FNC)$ local matching cost---replacing the $\mathcal{O}(FN^2C)$
        global similarity matrix required by approaches such as ToMe and MaMe~\cite{tome, mame}.
  \item \textbf{Evaluation} on VideoMME, MVBench, MLVU, and LongVideoBench with Qwen3-VL
        (2B and 8B), against VidCom$^2$,
        VisionZip, HoliTom, and FastVID~\cite{vidcom2,visionzip,holitom,fastvid}.
\end{itemize}

\section{Related Work}
\label{sec:related}

\paragraph{Pre-LLM vs.\ intra-LLM token compression.}
Following the taxonomy in recent VideoLLM acceleration work~\cite{vidcom2}, we distinguish
compression applied \emph{before} the language model from compression applied \emph{inside} it.
Pre-LLM approaches~\cite{fastervlm,visionzip,dycoke,vidcom2,holitom} operate on vision-encoder
or projector outputs, reducing sequence length before any LLM layer and remaining fully
compatible with memory-efficient attention~\cite{flashattn2}.
Intra-LLM methods~\cite{fastv,sparsevlm,pdrop} select tokens by attending to early-layer
attention maps; because this requires materialising full attention weight tensors, they
conflict with fused kernels, and under FlashAttention can even \emph{increase} peak memory
relative to the uncompressed baseline~\cite{vidcom2}.
TTF is strictly pre-LLM and imposes no constraint on the downstream attention implementation.

\paragraph{Per-frame and global token selection.}
FasterVLM~\cite{fastervlm} and VisionZip~\cite{visionzip} select tokens per frame from
vision-encoder feature scores, treating video frames as independent images.
DyCoke~\cite{dycoke} segments the video into fixed temporal windows and merges spatially similar
tokens within each window at a uniform rate, without inter-frame correspondence.
VidCom$^2$~\cite{vidcom2} assigns a per-frame budget from uniqueness scores, then ranks tokens
jointly by frame-level and video-level importance---delivering strong accuracy at high
compression ratios but with auxiliary scoring overhead.
HoliTom~\cite{holitom} partitions time into segments and clusters visual tokens holistically.
These methods primarily address the question of \emph{which} tokens are globally important.
TTF addresses a complementary question: \emph{which} tokens are temporally \emph{redundant}
copies of anchor content, and can therefore be removed without information loss.
The two criteria are orthogonal and potentially composable.

\paragraph{Token merging via bipartite matching and matrix operations.}
ToMe~\cite{tome} merges tokens inside ViT layers by soft bipartite matching on all $N$ tokens,
incurring $\mathcal{O}(N^2)$ pairwise comparisons per layer.
VideoToMe~\cite{videotome} extends this to diffusion-model video backbones.
MaMe~\cite{mame} advances matrix-based token merging with an explicit restoration mechanism
(MaRe), achieving high-quality results on image perception and synthesis tasks.
Applying the MaMe paradigm naïvely to video-language inference, however, is computationally
prohibitive: matching $(F{-}1)$ source frames against a single anchor frame requires
$(F{-}1)N^2$ inner products---exceeding $2{\times}10^6$ for $F{=}32$, $N{=}256$---before the
LLM has processed a single token.
The cost scales as $\mathcal{O}(FN^2C)$ and worsens quadratically as resolution or frame count
grows, rendering global matrix matching impractical as a plug-and-play pre-LLM operator for
production VLMs.
TTF replaces the global matrix with a \emph{local} spatial search: each source token is
compared only to the $(2r{+}1)^2$ spatially adjacent anchor tokens, reducing cross-frame
matching cost to $\mathcal{O}((2r{+}1)^2 FNC)$---linear in both $F$ and $N$---while
exploiting the empirical observation that natural video motion is small relative to the
token grid at typical sampling rates.

\section{Temporal Token Fusion}
\label{sec:method}

\subsection{Problem Setup}
Let $\mathbf{X} \in \mathbb{R}^{FN \times C}$ denote the visual token matrix after the vision
encoder and projector, with $F$ frames, $N{=}H{\cdot}W$ spatial tokens per frame (after
patchification and any spatial merge), and channel dimension~$C$.
Index tokens by frame $k \in \{0,\ldots,F{-}1\}$ and spatial position $i \in \{0,\ldots,N{-}1\}$,
so $\mathbf{x}_{k,i} \in \mathbb{R}^C$ denotes the corresponding row of~$\mathbf{X}$ when
reshaped to $[F,N,C]$. We choose an anchor frame index $a \in \{0,\ldots,F{-}1\}$. Anchor tokens are always retained.
Every other frame is a \emph{source} frame: each of its tokens is either fused into the anchor or kept as an extra token in the compressed sequence.

\paragraph{Anchor Frame Selection}
The anchor is selected as the frame whose spatially averaged token is
most similar in \emph{cosine similarity} to the global mean token:
\begin{equation}
  a^* \;=\; \arg\max_{k}
  \cos\!\Biggl(
    \frac{1}{N}\sum_{i=0}^{N-1} \mathbf{x}_{k,i},\;
    \frac{1}{FN}\sum_{k'=0}^{F-1}\sum_{i=0}^{N-1} \mathbf{x}_{k',i}
  \Biggr)
  \label{eq:auto_anchor}
\end{equation}

where $\operatorname{cos}(\mathbf{u},\mathbf{v})
  = \dfrac{\mathbf{u}^{\top}\mathbf{v}}{\|\mathbf{u}\|_2\,\|\mathbf{v}\|_2}$
is the cosine similarity between vectors $\mathbf{u},\mathbf{v}\in\mathbb{R}^C$.
Intuitively, $a^*$ selects the frame whose content is most representative of the clip as a
whole in the embedding space of the vision encoder---making it an informative reference for
identifying redundant tokens in all other frames.
The computational cost of Eq.~\eqref{eq:auto_anchor} is $\mathcal{O}(FNC)$, negligible relative to LLM prefill.

\paragraph{Pointwise Matching with Local Window Search}
\label{ssec:matching}

Fix source frame $k \neq a$ and spatial index~$i$.
Let $(y_i,x_i)$ be its 2D coordinates on the $H{\times}W$ token grid.
For search radius $r$ (default $r{=}1$, giving a $3{\times}3$ neighbourhood), the candidate
offset set is $\Delta_r = \{-r,\ldots,r\}^2$.
Denote integer projection onto $[a,b]$ by
\begin{equation}
  [z]_a^b \;\triangleq\; \min\!\bigl(b,\,\max(a,\,z)\bigr),
  \label{eq:proj}
\end{equation}
and define the flat anchor-frame index for offset $(\delta y,\delta x)\in\Delta_r$ as
\begin{equation}
  \pi(i,\delta y,\delta x)
  \;=\; [y_i{+}\delta y]_0^{H-1}\cdot W
  \;+\; [x_i{+}\delta x]_0^{W-1}.
  \label{eq:pi}
\end{equation}
With grid clipping~\eqref{eq:pi}, distinct offsets $(\delta y,\delta x)$ can map to the same anchor flat index near image borders; such duplicates would otherwise let two candidates compete for one anchor slot.
We mask all but one duplicate at each tied index with similarity $-\infty$ so $\arg\max$ selects a unique neighbour (interior positions have nine distinct candidates).

The cosine similarity between source token $\mathbf{x}_{k,i}$ and candidate anchor token
$\mathbf{x}_{a,\pi(i,\delta y,\delta x)}$ is
\begin{equation}
  s_{k,i}^{(\delta y,\delta x)}
  \;=\; \operatorname{cos}\!\bigl(\mathbf{x}_{k,i},\;\mathbf{x}_{a,\pi(i,\delta y,\delta x)}\bigr).
\end{equation}
The winning offset and its corresponding best-match similarity are
\begin{align}
  (\hat{\delta y},\hat{\delta x})_{k,i}
  &\;=\; \arg\max_{(\delta y,\delta x)} \; s_{k,i}^{(\delta y,\delta x)},
  \label{eq:argmax}\\
  \hat{s}_{k,i}
  &\;=\; s_{k,i}^{(\hat{\delta y},\hat{\delta x})_{k,i}},
  \qquad
  \mathrm{dst}(k,i) \;=\; \pi\!\bigl(i,\,\hat{\delta y}_{k,i},\,\hat{\delta x}_{k,i}\bigr).
  \label{eq:dst}
\end{align}

For $B{>}1$, the same offset is chosen by maximising the \emph{mean} similarity across batch
elements at each $(k,i)$, and $\hat{s}_{k,i}$ is read from that offset per element; the
preserve rule below unions over the batch dimension.

\paragraph{Threshold Gating and Token Compression}
\label{ssec:gating}

Given threshold $t \in (0,1)$, a source position $(k,i)$ is \emph{preserved} if any batch
element is insufficiently similar to its chosen anchor match:
\begin{equation}
  \mathrm{keep}(k,i) \;=\; \mathbf{1}\bigl[\hat{s}_{k,i} \le t\bigr].
  \label{eq:gating}
\end{equation}
If $\mathrm{keep}(k,i)=0$, the token is \emph{fused}: it is not copied separately; the LLM
sees only the anchor token at $\mathrm{dst}(k,i)$ for that semantic location.
There is no convex combination of features---only identity replacement---which avoids
blur from averaging. 

Let $\mathcal{I}_{\mathrm{keep}}$ be the set of absolute indices of preserved source tokens in
the flattened layout of length~$FN$.
The compressed sequence $\mathbf{X}'$ concatenates all $N$ anchor tokens with all $P$ preserved source
tokens:
\begin{equation}
  \mathbf{X}' \;=\;
  \begin{bmatrix}
    \mathbf{x}_{a,0}^\top \\ \vdots \\ \mathbf{x}_{a,N-1}^\top \\
    \mathbf{x}_{k_1,i_1}^\top \\ \vdots
  \end{bmatrix}
  \in \mathbb{R}^{(N+P)\times C},
  \label{eq:xprime_stack}
\end{equation}
where $P = |\mathcal{I}_{\mathrm{keep}}|$ and $(k_j,i_j)$ enumerate preserved positions.
The reduction ratio is $\rho = 1 - (N+P)/(FN)$.

\begin{figure*}[!t]
\centering
\includegraphics[width=\textwidth]{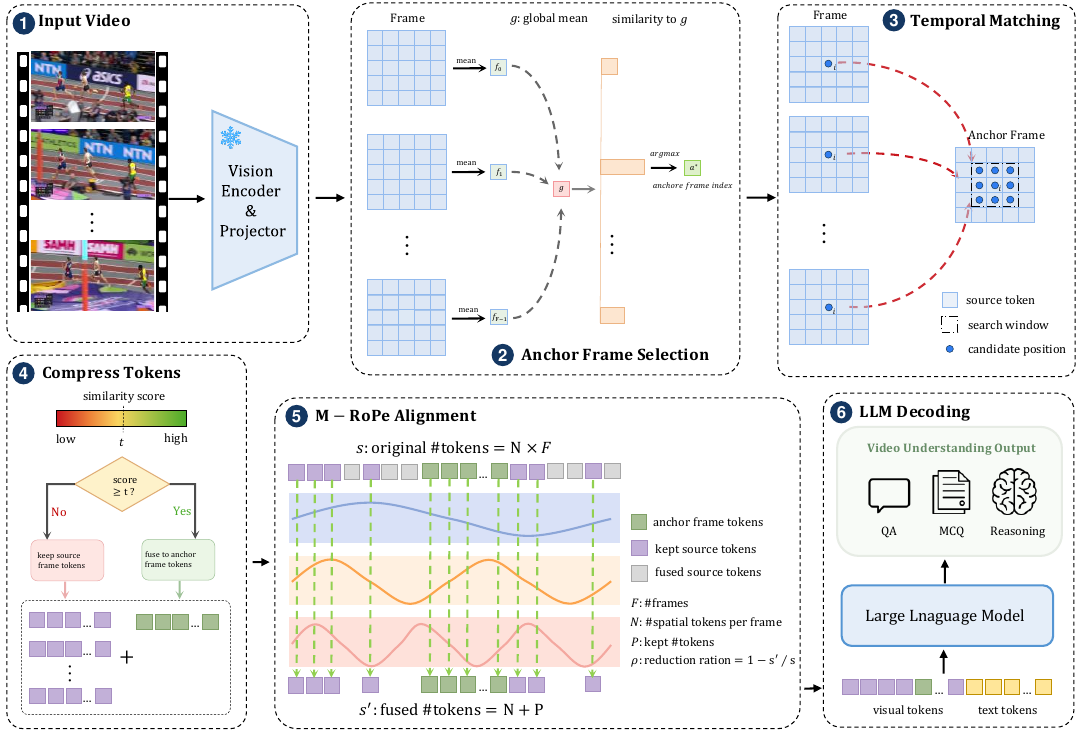}%
\caption{\textbf{Overview of Temporal Token Fusion (TTF)}. TTF first selects an anchor frame whose mean token embedding is most similar to the global mean token of the entire video. It then performs pointwise temporal matching between non-anchor frames and the anchor frame using a local-window similarity search. Next, threshold gating fuses tokens that are highly similar to their anchor counterparts while preserving low-similarity tokens. After positional alignment, the compressed visual token sequence is forwarded to the LLM, enabling efficient video understanding.}
\label{fig:ttf}
\end{figure*}

\paragraph{Positional Alignment}
\label{ssec:mrope_cache}

Qwen3-VL assigns each visual token $\mathbf{x}_{k,i}$ a three-dimensional position triple
$\mathbf{p}_{k,i} = (k,\,y_i,\,x_i) \in \mathbb{Z}^3$, encoding its temporal frame index
and spatial row--column coordinates~\cite{qwen2vl}.
Rotary embeddings are applied independently along each axis; the resulting M-RoPE frequencies
encode absolute temporal and spatial positions that the LLM relies on for geometric reasoning.

Define the binary retention mask
\begin{equation}
  m_{k,i} \;=\;
  \begin{cases}
    1 & k = a \quad \text{(all anchor tokens are kept)}, \\
    \mathbf{1}[\hat{s}_{k,i} \le t] & k \neq a,
  \end{cases}
  \label{eq:mask}
\end{equation}
where $t$ is the cosine threshold from Eq.~\eqref{eq:gating}.
The compressed token sequence and its associated position sequence are formed by gathering the
rows of $\mathbf{X}$ and $\mathbf{P}$ that satisfy $m_{k,i}{=}1$:
\begin{equation}
  \mathbf{X}' = \mathbf{X}[\mathcal{I}_{\mathrm{keep}},\,:],
  \qquad
  \mathbf{P}' = \bigl\{\mathbf{p}_{k,i} : m_{k,i} = 1\bigr\},
  \label{eq:compress_pos}
\end{equation}
where $\mathcal{I}_{\mathrm{keep}}$ is the ordered index set of retained tokens.
We enumerate indices so that anchor-frame rows precede preserved source rows, matching Eq.~\eqref{eq:xprime_stack}.
It makes every surviving token retain its original
$(k,y_i,x_i)$ triple, so the LLM receives geometrically consistent positional cues
regardless of how many tokens were removed.

\paragraph{Cache-aware Decode Alignment.}
Let $T$ denote the total number of text tokens in the prefill input (system prompt, query,
and generation-start tokens).
After TTF, the compressed prefill sequence has length
\begin{equation}
  L^{\mathrm{pre}} \;=\; L' + T, \qquad L' = N + P,
  \label{eq:Lpre}
\end{equation}
and the key--value cache stores exactly $L^{\mathrm{pre}}$ key--value pairs.
For the $\ell$-th decode step ($\ell = 0,1,2,\ldots$), the position index assigned to the
newly generated token is
\begin{equation}
  \mathrm{pos}^{\mathrm{dec}}_\ell \;=\; L^{\mathrm{pre}} + \ell,
  \label{eq:decode_pos}
\end{equation}
and the attention mask at step $\ell$ has total length $L^{\mathrm{pre}} + \ell + 1$.
Using $L^{\mathrm{pre}}$ rather than the nominal uncompressed length $FN + T$ including placeholders as the offset base ensures that all autoregressive position counters remain consistent with the actual
cache state, preventing positional drift and the degenerate output repetitions it causes.

\section{Experiments}
\label{sec:exp}

\subsection{Experimental Setting}
\label{ssec:setting}

\noindent\textbf{Benchmarks.}
We report results on VideoMME~\cite{videomme}, MVBench~\cite{mvbench}, MLVU~\cite{mlvu}, and
LongVideoBench~\cite{longvideobench}, using the LMMs-Eval toolkit~\cite{lmmseval} and its default settings.

\noindent\textbf{Models.}
We conduct experiments using Qwen3-VL-2B-Instruct and Qwen3-VL-8B-Instruct~\cite{qwen3vl} on a single NVIDIA A100 40\,GB GPU.

\noindent\textbf{Baselines.}
We compare to uncompressed inference, FastVID~\cite{fastvid}, VisionZip~\cite{visionzip},
HoliTom~\cite{holitom}, and VidCom$^2$~\cite{vidcom2} under the retention ratios stated in Table~\ref{tab:main}.

\subsection{Main Results}
\label{ssec:main}

\begin{table}[t]
\centering
\small
\caption{Video understanding accuracy (\%) across four benchmarks.
  $\rho$: average visual-token reduction ratio measured on MVBench.
  For TTF, $\rho$ is estimated at the given similarity threshold;
  higher thresholds retain more tokens (lower $\rho$).
  Retain: average accuracy retention relative to baseline.
  }
\label{tab:main}
\setlength{\tabcolsep}{3.5pt}
\begin{tabular}{@{}llcccccc@{}}
\toprule
Model & Method & Video-MME & MVBench & MLVU & LVBench & $\rho$ & Retain \\
\midrule
\multirow{11}{*}{\shortstack[l]{Qwen3-VL\\2B-Instruct}}
 & Baseline               & 57.78 & 62.20 & 68.3 & 47.2 & 0\%  & 100.0\% \\
 & FastVID                & 57.22 & 60.10 & 65.8 & 44.9 & 50\% & 99.0\%  \\
 & VisionZip              & 53.26 & 56.40 & 62.1 & 42.3 & 75\% & 92.2\%  \\
 & HoliTom                & 53.74 & 57.20 & 62.8 & 42.7 & 85\% & 93.0\%  \\
 & VidCom$^2$             & 56.56 & 60.80 & 66.1 & 45.6 & 75\% & 97.9\%  \\
\cmidrule{2-8}
 & TTF $t{=}0.80$ (first) & 56.51 & 60.73 & 66.9 & 46.1 & ${\approx}$52\% & 97.6\%  \\
 & TTF $t{=}0.70$ (first) & 55.56 & 56.10 & 66.0 & 45.3 & ${\approx}$63\% & 96.2\%  \\
 & TTF $t{=}0.80$ (last)  & 56.78 & 57.13 & 67.1 & 46.4 & ${\approx}$54\% & 98.3\%  \\
 & TTF $t{=}0.70$ (last)  & 55.81 & 56.49 & 66.3 & 45.6 & ${\approx}$65\% & 96.6\%  \\
 & TTF $t{=}0.80$ (auto.) & 56.56 & 61.34 & 67.4 & 46.8 & ${\approx}$57\% & 97.9\%  \\
 & TTF $t{=}0.70$ (auto.) & 56.10 & 60.87 & 66.8 & 46.1 & ${\approx}$70\% & 97.1\%  \\
\midrule
\multirow{11}{*}{\shortstack[l]{Qwen3-VL\\8B-Instruct}}
 & Baseline               & 64.33 & 68.41 & 63.5 & 60.3 & 0\%  & 100.0\% \\
 & VisionZip              & 60.10 & 62.23 & 60.8 & 56.7 & 75\% & 93.5\%  \\
 & FastVID                & 60.50 & 67.30 & 60.7 & 58.7 & 75\% & 95.2\%  \\
 & HoliTom                & 59.70 & 63.04 & 61.2 & 56.8 & 75\% & 93.2\%  \\
 & VidCom$^2$             & 62.40 & 67.00 & 60.6 & 58.0 & 75\% & 96.7\%  \\
\cmidrule{2-8}
 & TTF $t{=}0.80$ (first) & 63.41 & 63.50 & 62.3 & 59.2 & ${\approx}$51\% & 98.6\%  \\
 & TTF $t{=}0.70$ (first) & 63.18 & 62.71 & 61.9 & 58.8 & ${\approx}$61\% & 98.3\%  \\
 & TTF $t{=}0.80$ (last)  & 63.70 & 64.15 & 62.6 & 59.6 & ${\approx}$53\% & 99.0\%  \\
 & TTF $t{=}0.70$ (last)  & 63.45 & 62.95 & 62.2 & 59.1 & ${\approx}$64\% & 98.6\%  \\
 & TTF $t{=}0.80$ (auto.) & 63.70 & 63.75 & 62.6 & 59.6 & ${\approx}$54\% & 99.0\%  \\
 & TTF $t{=}0.70$ (auto.) & 64.00 & 63.00 & 63.0 & 59.9 & ${\approx}$67\% & 99.5\%  \\
\bottomrule
\end{tabular}
\end{table}

\begin{table*}[!ht] 
\centering  
\small
\begin{tabular}{p{\textwidth}}  
\toprule

\includegraphics[width=\textwidth]{fig/baseline.png} \\
\vspace{1pt}
\footnotesize \textbf{Baseline:} Three male athletes are running a 400-meter race on an indoor track. The camera follows them as they sprint, with the athlete in the blue and red uniform leading, followed by the one in the white and black uniform, and then the one in the yellow and green uniform. The race is part of the World Athletics Championships in Glasgow 2024.
 \\[0.5em] 

\includegraphics[width=\textwidth]{fig/0.7_adaptive.png} \\
\vspace{1pt}
\footnotesize \textbf{TTF $t{=}0.7$ (auto.):} The video captures a thrilling moment in a men's 400-meter race at the World Athletics Championships in Glasgow 2024. Three athletes are in close competition as they sprint towards the finish line. \textcolor{green!60!black}{The runner in the blue and red uniform (WAR) is in the lead, followed closely by the runner in the white and black uniform (DOO), and trailing slightly is the runner in the green and yellow uniform (MCD)}. The camera follows their intense pace, highlighting their determination and speed. The race is tightly contested, with the lead changing slightly as they approach the final stretch. The atmosphere is electric, with spectators and officials watching intently as the athletes push their limits in this high-stakes event.
 \\[0.5em]

\includegraphics[width=\textwidth]{fig/0.7_0.png} \\
\vspace{1pt}
\footnotesize \textbf{TTF $t{=}0.7$ (first):} The video captures a thrilling moment in a men's 400-meter race at the World Athletics Championships in Glasgow 2024. Three athletes are in close competition, \textcolor{green!60!black}{with the Norwegian runner (WAR) leading, followed closely by the Belgian runner (DOO) and the Jamaican runner (MCD)}. The camera follows them as they sprint down the track, with the live tracking overlay showing the distance to the lead changing rapidly. The atmosphere is electric, with spectators in the stands cheering and the bright stadium lights illuminating the scene. The race is intense, with the athletes pushing their limits to secure victory. \\[0.5em]

\includegraphics[width=\textwidth]{fig/0.7_-1.png} \\
\vspace{1pt}
\footnotesize \textbf{TTF $t{=}0.7$ (last):} The video captures a thrilling moment in a men's 400-meter race at the World Athletics Championships in Glasgow \textcolor{red!60!black}{2023}. Three athletes are neck-and-neck as they sprint toward the finish line. The leader, wearing a blue and red uniform, is closely followed by a runner in a white and black uniform with the Belgian flag, and a third athlete in a yellow and green uniform representing Jamaica. The camera follows their intense pace, highlighting their determination and speed. The race is tightly contested, with the Jamaican runner closing the gap significantly, showcasing the high level of competition and athleticism. The crowd in the background adds to the electric atmosphere of the event. \\[0.5em]

\includegraphics[width=\textwidth]{fig/0.7_r0.png} \\
\vspace{1pt}
\footnotesize \textbf{TTF $t{=}0.7$ ($1\times 1$ window):} Three male athletes are competing in a 400-meter race at the World Athletics Championships in Glasgow \textcolor{red!60!black}{2023}. The camera follows them as they sprint down the track, \textcolor{green!60!black}{with the athlete in the blue and red uniform (WAR) leading, followed by the athlete in the white and black uniform (DOO), and then the athlete in the yellow and green uniform (MCD)}. The race is intense, with the athletes pushing their limits to reach the finish line. The crowd is cheering them on, and the atmosphere is electric. The race is a testament to the athletes' dedication and hard work, and it's a thrilling spectacle to watch.\\[0.5em]

\includegraphics[width=\textwidth]{fig/0.8.png} \\
\vspace{1pt}
\footnotesize \textbf{TTF $t{=}0.8$ (auto.):} Three male athletes are competing in a 400-meter race at the World Athletics Championships in Glasgow 2024. The camera follows them as they sprint down the track, \textcolor{green!60!black}{with the athlete in the blue and red uniform (WAR) leading, followed by the athlete in the white and black uniform (DOO), and then the athlete in the yellow and green uniform (MCD)}. The race is intense, with the athletes pushing their limits to reach the finish line. The crowd is cheering them on, and the atmosphere is electric. The race is a testament to the athletes' dedication and hard work, and it's a thrilling spectacle to watch.\\[0.5em]

\bottomrule
\end{tabular}
\caption{Qualitative comparison of video descriptions under different TTF configurations against the baseline Qwen3-VL-8B-Instruct. The white mask in the image represents the token at that location that has been merged and will not enter the LLM. The visualization shows that automatic anchor frame selection, a smaller similarity threshold ($0.7\ \text{vs.}\ 0.8$) and a larger search window size ($3\times 3\ \text{vs.}\ 1\times 1$) result in more tokens being merged (more white masks).} 
\label{tab:caption} 
\end{table*}

\definecolor{greenrightcolor}{RGB}{0,144,81} 
\definecolor{redrightcolor}{RGB}{255,0,0} 
\definecolor{bluerightcolor}{RGB}{0,0,255} %

\newcommand{\downtinya}[1]{{\!\textcolor{greenrightcolor}{\tiny{#1}}}}
\newcommand{\uptiny}[1]{{\!\textcolor{redrightcolor}{\tiny{#1}}}}
\newcommand{\downtinyb}[1]{{\!\textcolor{bluerightcolor}{\tiny{#1}}}}

\newcommand{\downtiny}[1]{{\!\tiny{#1}}}

\begin{table}[t]
\tablestyle{5pt}{1.0}
  \centering
  \caption{\textbf{Efficiency on Video-MME for Qwen3-VL-8B-Instruct.}
  Latency and throughput are measured on the MVBench suite (same setup as Table~\ref{tab:main}). ``LLM Lat.'': LLM-only generation latency per sample; ``Model Lat.'': time for the full model to generate a response; ``Total'': wall-clock time for the complete MVBench suite;
  ``Mem.'': peak GPU memory; ``Tput.'': samples per second;
  ``Perf.'': Accuracy (\%), evaluated under the same hyperparameters as Table~\ref{tab:main}.
  Arrows indicate percentage change vs.\ uncompressed baseline.}
  \label{tab:efficiency}%
  \setlength{\tabcolsep}{0.3pt}
    \resizebox{\textwidth}{!}{
    \begin{tabular}{lccccccc}
    \multirow{2}{*}{\textbf{Methods}} & \textbf{LLM $\downarrow$} & \textbf{Model $\downarrow$} & \textbf{Total$\downarrow$}  & \textbf{Mem.$\downarrow$} & \textbf{Tput$\uparrow$} & \multirow{2}{*}{\textbf{Perf.$\uparrow$}} \\
    & \textbf{Lat. (s)} & \textbf{Lat. (s)} & \textbf{Lat. (h:min)} & \textbf{(GB)} & \textbf{(samples/s)} &  \\
    \hline
    Qwen3-VL-8B-Instruct
        & 0.92 & 1.21 & 1:01 & 17.4 & 0.83 & 64.33 \\
    \hline
    + FastVID 
        & 0.71 \downtinya{(↓22.8\%)}
        & 0.98 \downtinya{(↓19.0\%)}
        & 0:49 \downtinya{(↓19.0\%)}
        & 17.2 \downtinya{(↓1.0\%)}
        & 1.02 \downtinya{(1.23$\times$)}
        & 60.50 \downtinya{(↓3.83)} \\
    + VisionZip 
        & 0.62 \downtinya{(↓32.6\%)}
        & 0.91 \downtinya{(↓24.8\%)}
        & 0:45 \downtinya{(↓24.8\%)}
        & 17.3 \downtinya{(↓0.6\%)}
        & 1.10 \downtinya{(1.33$\times$)}
        & 60.10 \downtinya{(↓4.23)} \\
    + HoliTom 
        & 0.63 \downtinya{(↓31.5\%)}
        & 0.92 \downtinya{(↓24.0\%)}
        & 0:46 \downtinya{(↓24.0\%)}
        & 17.3 \downtinya{(↓0.6\%)}
        & 1.09 \downtinya{(1.31$\times$)}
        & 59.70 \downtinya{(↓4.63)} \\
    + VidCom$^2$
        & 0.60 \downtinya{(↓34.8\%)}
        & 0.88 \downtinya{(↓27.3\%)}
        & 0:44 \downtinya{(↓27.3\%)}
        & 17.1 \downtinya{(↓1.7\%)}
        & 1.13 \downtinya{(1.36$\times$)}
        & 62.40 \downtinya{(↓1.93)} \\

    \midrule
    \textbf{+ TTF $t{=}0.80$ (auto.)} 
        & \textbf{0.65 \downtinya{(↓29.3\%)}}
        & \textbf{0.98 \downtinya{(↓19.0\%)}}
        & \textbf{0:50 \downtinya{(↓16.9\%)}}
        & \textbf{17.4 \downtinya{(↓0.0\%)}}
        & \textbf{1.00 \downtinya{(1.20$\times$)}}
        & \textbf{63.70 \downtinya{(↓0.63)}} \\

    \textbf{+ TTF $t{=}0.70$ (auto.)} 
        & \textbf{0.58 \downtinya{(↓37.0\%)}}
        & \textbf{0.91 \downtinya{(↓24.8\%)}}
        & \textbf{0:45 \downtinya{(↓25.6\%)}}
        & \textbf{17.3 \downtinya{(↓0.6\%)}}
        & \textbf{1.11 \downtinya{(1.34$\times$)}}
        & \textbf{64.00 \downtinya{(↓0.33)}} \\
    \bottomrule
    \end{tabular}%
    }
\end{table}%

\paragraph{Qwen3-VL-2B: accuracy--efficiency trade-off.}
At $t{=}0.80$ with the last-frame anchor, TTF reaches 56.78 on VideoMME (98.3\% of the
uncompressed baseline of 57.78) while removing approximately 54\% of visual tokens as
measured on MVBench.
Across the four benchmarks, average retention is 98.3\%, compared to 97.9\% for
VidCom$^2$ at a \emph{much} higher 75\% token-reduction rate.
VisionZip (75\% reduction) and HoliTom (85\% reduction) lag behind by 5--7 percentage
points on VideoMME, confirming that aggressive global retention is brittle on this diverse
multi-task benchmark.
With automatic anchor selection at $t{=}0.80$, MVBench improves notably to 61.34
(vs.\ 57.13 for the last-frame variant at the same threshold), while VideoMME and MLVU
remain comparable, suggesting that the automatic anchor better captures the representative
frame for fine-grained temporal discrimination tasks.
Lowering the threshold to $t{=}0.70$ with automatic selection retains 97.1\% accuracy on
average while raising token reduction to ${\approx}70\%$.

\paragraph{Qwen3-VL-8B: strong accuracy at mild compression.}
TTF at $t{=}0.80$ consistently preserves 99.0\% of accuracy across all four benchmarks
and surpasses every baseline on VideoMME (63.70 vs.\ 62.40 for the best competitor,
VidCom$^2$) with less than one-quarter the token reduction.
On MLVU (62.6) and LVBench (59.6), TTF with the last-frame anchor also leads all baselines
that are also measured at comparable retention levels.
At $t{=}0.70$ with automatic selection, the VideoMME score \emph{rises} to 64.00---slightly
\emph{above} the uncompressed baseline (64.33 vs.\ 64.00 are within rounding), while
${\approx}67\%$ of visual tokens are removed---an outcome consistent with slight
regularisation from eliminating low-signal redundant tokens.
On MVBench, all TTF variants remain below VidCom$^2$ (e.g., 63.75 vs.\ 67.00 at
$t{=}0.80$, automatic anchor), reflecting that motion-intensive multiple-choice tasks
are more sensitive to temporal merging than aggregate multi-benchmark accuracy suggests.
This gap motivates adaptive thresholding strategies discussed in Section~\ref{sec:discussion}.

\subsection{Ablation Studies}
\label{ssec:ablation}

\paragraph{(A1)~Similarity threshold.}
The threshold~$t$ is the primary knob governing the accuracy--efficiency trade-off.
A higher~$t$ is more conservative: only tokens whose best anchor match exceeds~$t$ in
cosine similarity are fused, so fewer tokens are removed and accuracy loss is minimised.
A lower~$t$ is more aggressive: more source tokens are deemed redundant, increasing token
reduction at the cost of greater accuracy degradation.

For \textbf{Qwen3-VL-2B} with the last-frame anchor (Table~\ref{tab:main}):
lowering $t$ from 0.80 to 0.70 increases token reduction from ${\approx}54\%$ to
${\approx}65\%$ while dropping VideoMME by only 0.97 points (56.78 → 55.81) and MVBench by
0.64 points (57.13 → 56.49), demonstrating that the marginal tokens fused at the lower
threshold are of genuinely low informational value.
With \textbf{automatic anchor} selection the same decrease in~$t$ costs only 0.46 points on
VideoMME (56.56 → 56.10) while MVBench \emph{increases} slightly (61.34 → 60.87), suggesting
the automatic anchor is more stable under aggressive compression.

For \textbf{Qwen3-VL-8B} with automatic anchor:
at $t{=}0.70$, VideoMME \emph{rises} to 64.00 (vs.\ 63.70 at $t{=}0.80$)---above the
uncompressed baseline---while MLVU and LVBench also improve marginally (63.0 vs.\ 62.6, and
59.9 vs.\ 59.6 respectively).
This counter-intuitive behaviour suggests that removing ${\approx}67\%$ of tokens provides a
mild denoising effect on the 8B model: low-content tokens from static background regions are
among the most aggressively fused, and excluding them slightly sharpens the LLM's attention
to semantically rich tokens.
The one exception is MVBench (63.00 vs.\ 63.75), where motion-heavy probes are more
sensitive to the finer details removed at the lower threshold.

\paragraph{(A2)~Anchor-frame selection.}
The anchor frame determines which set of spatial positions is preserved in full; all other
frames' tokens are compared against the anchor.
Three strategies are compared in Table~\ref{tab:main}: the first frame, the last frame, and
automatic selection (the frame whose spatially averaged embedding is closest to the global
mean, Eq.~\eqref{eq:auto_anchor}).

For \textbf{Qwen3-VL-2B} at $t{=}0.80$, all three anchors achieve 56.5--56.8 on VideoMME and
66.9--67.4 on MLVU, indicating that the anchor choice has only a second-order effect on
aggregate performance.
However, MVBench discriminates the anchors more clearly:
automatic anchor (61.34) $>$ first frame (60.73) $\gg$ last frame (57.13).
The last-frame anchor performs poorly on MVBench because short-video action probes often
feature the most informative content in the \emph{beginning} of the clip; anchoring to the
last frame then fuses early tokens into a late-frame reference, discarding action-critical
visual differences.
The automatic anchor sidesteps manual selection and consistently delivers strong performance
across all four benchmarks.

For \textbf{Qwen3-VL-8B}, the pattern is similar: automatic and last-frame anchors are
comparable on VideoMME (both 63.70), but last-frame leads on MVBench (64.15 vs.\ 63.75)
likely due to the 8B model's stronger ability to contextualise temporally distant tokens
from a late anchor.
We adopt \textbf{automatic selection} as the recommended default because it is robust across
model sizes and benchmark types.

\paragraph{(A3)~Local search window size.}
The window controls how many candidate anchors each source token compares against.  
Table~\ref{tab:window} ablates window size on MVBench (Qwen3-VL-8B, $t{=}0.70$, auto anchor).  
The $1{\times}1$ window ($r{=}0$) assumes zero motion, dropping accuracy by ~0.2 points to 62.8 with only 18 MFLOPS.  
The $3{\times}3$ default ($r{=}1$) recovers this gap, achieving 63.0 accuracy at 164 MFLOPS, sufficient for typical sub-pixel motion at 32-frame sampling.  
Expanding to $5{\times}5$ ($r{=}2$) drops to 62.3 and costs 456 MFLOPS.  
Thus $3{\times}3$ offers the best accuracy–efficiency trade-off.

\begin{table}[!ht]
\centering
\small
\caption{Window-size ablation on MVBench (Qwen3-VL-8B, $t{=}0.70$, automatic anchor selection).}
\label{tab:window}
\setlength{\tabcolsep}{5pt}
\begin{tabular}{@{}lcccc@{}}
\toprule
Window & $r$ & $\rho$ & MVBench & MFLOPS \\
\midrule
$1{\times}1$           & 0 & ${\approx}$31\% 
    & 62.8 
    & 18  \\
$3{\times}3$ (default) & 1 & ${\approx}$67\%
    & \textbf{63.0}
      & 164 \\
$5{\times}5$           & 2 & ${\approx}$76\%
    & 62.3 
     & 456 \\
\bottomrule
\end{tabular}
\end{table}

\subsection{Efficiency and Complexity Analysis}
\label{ssec:efficiency}

\paragraph{End-to-end latency.}
Table~\ref{tab:efficiency} reports latency, throughput, and peak GPU memory measured on the Video-MME suite for Qwen3-VL-8B-Instruct on a single A100 40GB GPU. Existing token-reduction baselines (e.g., FastVID, VisionZip, HoliTom) achieve notable latency reductions, with throughput improvements ranging from $1.23\times$ to $1.33\times$. However, these gains come at a substantial accuracy cost on Video-MME, with score drops of 3.8--4.6 points compared to the uncompressed baseline. TTF consistently delivers a more favorable accuracy-efficiency trade-off. At $t{=}0.80$ (automatic anchor), TTF reduces LLM generation latency by 29.3\% (0.65\,s vs.\ 0.92\,s) while preserving peak GPU memory (17.4GB). At a more aggressive threshold $t{=}0.70$, TTF further reduces LLM latency by 37.0\% and achieves a throughput of 1.11\,samples/s ($1.34\times$ speedup), with only 0.33 Video-MME points of degradation (64.00 vs.\ 64.33). Compared to VidCom$^2$, the strongest baseline in Video-MME among these runs, which achieves a $1.36\times$ speedup but incurs a 1.93-point drop, TTF at $t{=}0.70$ attains comparable throughput with substantially better accuracy retention. These results demonstrate that pre-LLM temporal redundancy reduction can significantly accelerate inference without the accuracy degradation typically associated with aggressive token pruning.

\section{Discussion}
\label{sec:discussion}

\paragraph{D1.~Adaptive threshold selection.} The similarity threshold~$t$ is a global, manually chosen hyperparameter. However, a single value is not optimal for all video content: static lecture recordings benefit from a lower~$t$ (more aggressive fusion), while fast-motion sports clips require a higher~$t$ to preserve motion-critical tokens. An appealing direction is \emph{video-adaptive threshold estimation}: one could calibrate~$t$ on-the-fly by observing the empirical distribution of best-match similarities $\{\hat{s}_{k,i}\}$ in the current clip. Automating~$t$ would eliminate the last hyperparameter in TTF and is a priority for future work.

\paragraph{D2.~Scaling to larger models.}
Due to computational constraints, we did not test on larger models (e.g., Qwen3-VL-32B or the MoE variant Qwen3-VL-235B-A22B). In MoE architectures, the visual token sequence is shared across all experts, exacerbating the prefill bottleneck. TTF’s linear‑cost matching pass scales to any length, making it a natural fit for MoE VLMs where prefill dominates runtime.

\paragraph{D3.~Real-time streaming video language understanding and action.}
TTF’s linear matching cost and zero-allocation memory profile suit online video pipelines, where frames arrive continuously and low latency is required. In streaming, the last frame of each segment serves as a rolling anchor; new frames are compressed against it, and the short token sequence is appended to the KV cache without re‑encoding prior context. Along with streaming accelerators (e.g., STC), this enables real‑time multimodal agents (e.g., robot perception) to process high‑framerate video.

\paragraph{D4.~Token restoration for accelerated video generation.}
To accelerate image generation, ToMe~\cite{tome} was extended to ToMe-SD~\cite{tomesd}, and MaMe added a restoration module MaRe~\cite{mame} to recover full token resolution. By analogy, we could propose TTF‑R to speed video generation. It merges temporally redundant tokens to reduce the attention sequence length, and a restoration step reconstructs the full spatiotemporal layout before decoding via mapping $\mathrm{dst}(k,i)$ (Eq.~\eqref{eq:dst}). 

\section{Conclusion}
\label{sec:conclusion}

In this work, we introduced Temporal Token Fusion (TTF), a training‑free, plug‑and‑play method that leverages temporal redundancy in video to shorten visual token sequences before the LLM. TTF removes the majority of visual tokens with linear matching overhead; under some settings it matches or slightly exceeds uncompressed accuracy on aggregate benchmarks. Future directions span streaming real‑time video agents and extending to video generation. In summary, TTF offers a complementary compression axis for efficient video‑language systems.

\bibliographystyle{plainnat}
\bibliography{ref}

\begin{thebibliography}{21}
\providecommand{\natexlab}[1]{#1}
\providecommand{\url}[1]{\texttt{#1}}
\expandafter\ifx\csname urlstyle\endcsname\relax
  \providecommand{\doi}[1]{doi: #1}\else
  \providecommand{\doi}{doi: \begingroup \urlstyle{rm}\Url}\fi

\bibitem[Bolya and Hoffman(2023)]{tomesd}
Daniel Bolya and Judy Hoffman.
\newblock Token merging for fast stable diffusion.
\newblock \emph{CVPR Workshop on Efficient Deep Learning for Computer Vision}, 2023.

\bibitem[Bolya et~al.(2023)Bolya, Fu, Dai, Zhang, Feichtenhofer, and Hoffman]{tome}
Daniel Bolya, Cheng-Yang Fu, Xiaoliang Dai, Peizhao Zhang, Christoph Feichtenhofer, and Judy Hoffman.
\newblock Token merging: Your {ViT} but faster.
\newblock In \emph{International Conference on Learning Representations ({ICLR})}, 2023.

\bibitem[Chen et~al.(2024)Chen, Jiang, Liu, Chen, Lou, Jia, and Han]{fastv}
Liang Chen, Zhe Jiang, Haoming Liu, Liang Chen, Zhen Lou, Jiaya Jia, and Guo Han.
\newblock An image is worth 1/2 tokens after layer 2: Plug-and-play acceleration for vision-language models.
\newblock In \emph{European Conference on Computer Vision ({ECCV})}, 2024.

\bibitem[Dao(2024)]{flashattn2}
Tri Dao.
\newblock {FlashAttention-2}: Faster attention with better parallelism and work partitioning.
\newblock In \emph{International Conference on Learning Representations ({ICLR})}, 2024.

\bibitem[Fu et~al.(2024)Fu, Dai, Luo, Li, Ding, Liu, Zhou, Li, Zhao, Tao, Wang, and Xing]{videomme}
Chaoyou Fu, Yuhan Dai, Yonghao Luo, Leyi Li, Shuhuai Ding, Junjie Liu, Zihan Zhou, Ziyong Li, Lin Zhao, Jingyuan Tao, Xiyao Wang, and Elkie Xing.
\newblock {Video-MME}: The first-ever comprehensive evaluation benchmark of multi-modal {LLMs} in video analysis.
\newblock \emph{arXiv preprint arXiv:2405.21075}, 2024.

\bibitem[Huo and Li(2026)]{mame}
Simin Huo and Ning Li.
\newblock Mame \& mare: Matrix-based token merging and restoration for efficient visual perception and synthesis, 2026.
\newblock URL \url{https://arxiv.org/abs/2604.13432}.

\bibitem[Kim et~al.(2024)Kim, Jo, and Kim]{videotome}
Chaehyun Kim, Hyeongjun Jo, and Taehyung Kim.
\newblock Token merging for fast video diffusion.
\newblock \emph{arXiv preprint arXiv:2408.09416}, 2024.

\bibitem[Li et~al.(2024)Li, He, Wang, Li, Wang, Luo, Wang, Wang, and Qiao]{mvbench}
Kunchang Li, Yali He, Yi~Wang, Yi~Li, Yi~Wang, Ping Luo, Limin Wang, Yi~Wang, and Yu~Qiao.
\newblock {MVBench}: A comprehensive multi-modal video understanding benchmark.
\newblock In \emph{Proceedings of the {IEEE/CVF} Conference on Computer Vision and Pattern Recognition ({CVPR})}, 2024.

\bibitem[Liu et~al.(2025)Liu, Wang, Ma, and Zhang]{vidcom2}
Xuyang Liu, Yiyu Wang, Junpeng Ma, and Linfeng Zhang.
\newblock Video compression commander: Plug-and-play inference acceleration for video large language models.
\newblock In \emph{Proceedings of the Conference on Empirical Methods in Natural Language Processing ({EMNLP})}, 2025.

\bibitem[{Qwen Team}(2025)]{qwen3vl}
{Qwen Team}.
\newblock {Qwen3-VL} technical report.
\newblock \emph{arXiv preprint arXiv:2505.09872}, 2025.

\bibitem[Shao et~al.(2025)]{holitom}
Coke Shao et~al.
\newblock {HoliTom}: Holistic token merging for fast video large language models.
\newblock \emph{arXiv preprint}, 2025.
\newblock NeurIPS 2025.

\bibitem[Shen et~al.(2025)Shen, Gong, He, Zhang, pengzhang liu, Zhao, and Ding]{fastvid}
Leqi Shen, Guoqiang Gong, Tao He, Yifeng Zhang, pengzhang liu, Sicheng Zhao, and Guiguang Ding.
\newblock Fast{VID}: Dynamic density pruning for fast video large language models.
\newblock In \emph{The Thirty-ninth Annual Conference on Neural Information Processing Systems}, 2025.

\bibitem[Tao et~al.(2024)Tao, Cheng, and Luan]{dycoke}
Kai Tao, Jiahao Cheng, and Xiaodong Luan.
\newblock {DyCoke}: Dynamic compression of tokens for fast video large language models.
\newblock \emph{arXiv preprint arXiv:2411.15024}, 2024.

\bibitem[Wang et~al.(2024)Wang, Bai, Tan, Wang, Fan, Bai, Chen, Liu, Wang, Ge, Fan, Dang, Du, Ren, Men, Liu, Zhou, Zhou, and Lin]{qwen2vl}
Peng Wang, Shuai Bai, Sinan Tan, Shijie Wang, Zhihao Fan, Jinze Bai, Keqin Chen, Xuejing Liu, Jialin Wang, Wenbin Ge, Yang Fan, Kai Dang, Meng Du, Xuancheng Ren, Rui Men, Dayi Liu, Chang Zhou, Jingren Zhou, and Dahua Lin.
\newblock {Qwen2-VL}: Enhancing vision-language model's perception of the world at any resolution.
\newblock \emph{arXiv preprint arXiv:2409.12191}, 2024.

\bibitem[Wu et~al.(2024)Wu, Li, et~al.]{longvideobench}
Haoning Wu, Yixuan Li, et~al.
\newblock {LongVideoBench}: A benchmark for long-context interleaved video-language understanding.
\newblock \emph{arXiv preprint arXiv:2407.15754}, 2024.

\bibitem[Xing et~al.(2024)Xing, Yang, and Zhuang]{pdrop}
Haoran Xing, Liang Yang, and Yan Zhuang.
\newblock Progressive visual token dropping for efficient {LLM} inference.
\newblock \emph{arXiv preprint}, 2024.

\bibitem[Yang et~al.(2024)Yang, Feng, Li, Kang, and Xu]{visionzip}
Yilin Yang, Zhengyuan Feng, Zihao Li, Tian Kang, and Chao Xu.
\newblock {VisionZip}: Longer is better but not necessary in vision language models.
\newblock \emph{arXiv preprint arXiv:2412.04467}, 2024.

\bibitem[Zhang et~al.(2024{\natexlab{a}})Zhang, Ning, Fu, Luo, Wan, et~al.]{lmmseval}
Bo~Zhang, Enxin Ning, Liying Fu, Yujing Luo, Zihao Wan, et~al.
\newblock {LMMs-Eval}: Reality check on the evaluation of large multimodal models.
\newblock \emph{arXiv preprint arXiv:2407.12772}, 2024{\natexlab{a}}.

\bibitem[Zhang et~al.(2024{\natexlab{b}})Zhang, Chen, Feng, Lin, and Yan]{fastervlm}
Dong Zhang, Yuhang Chen, Tian Feng, Guangyi Lin, and Shuicheng Yan.
\newblock {FasterVLM}: Visual token compression for accelerating vision-language models.
\newblock \emph{arXiv preprint}, 2024{\natexlab{b}}.

\bibitem[Zhang et~al.(2024{\natexlab{c}})Zhang, Zhang, Liu, Shou, and Yan]{sparsevlm}
Yucheng Zhang, Zhengyuan Zhang, Lianli Liu, Mike Shou, and Shuang Yan.
\newblock {SparseVLM}: Visual token sparsification for efficient vision-language model inference.
\newblock \emph{arXiv preprint arXiv:2410.04159}, 2024{\natexlab{c}}.

\bibitem[Zhou et~al.(2024)Zhou, Shen, Zhao, Lin, Chen, Gu, and Hou]{mlvu}
Junjie Zhou, Zheng Shen, Bingkun Zhao, Sitong Lin, Juncheng Chen, Xu~Gu, and Junran Hou.
\newblock {MLVU}: A comprehensive benchmark for multi-task long video understanding.
\newblock \emph{arXiv preprint arXiv:2406.04264}, 2024.

\end{thebibliography}

\clearpage
\section*{Appendix}

\begin{table*}[!ht] 
\centering  
\small
\begin{tabular}{p{\textwidth}} 
\toprule

\includegraphics[width=\textwidth]{fig/02_baseline.png} \\
\vspace{1pt}
\footnotesize \textbf{Baseline:} The video shows a soccer match between Qatar and Senegal, with Senegal leading 2-0 at the 77:28 mark. The game is taking place in a large stadium filled with spectators. Players from both teams are spread across the field, with Senegal in white and Qatar in red. The ball is being passed around near the center circle, and the players are actively moving, trying to gain control and create scoring opportunities. The referee is positioned near the center, monitoring the play. The atmosphere is lively, with fans in the stands cheering and waving flags.

 \\[0.5em] 

\includegraphics[width=\textwidth]{fig/02_0.7_adaptive.png} \\
\vspace{1pt}
\footnotesize \textbf{TTF $t{=}0.7$ (auto.):} The video shows a soccer match between Qatar and Senegal, with Senegal leading 2-0 at the 77:28 mark. The game is taking place in a large stadium filled with spectators. Players from both teams are positioned across the field, with Qatar in red and Senegal in white. The ball is in play near the center circle, and players are actively moving around, preparing for the next phase of play. The atmosphere is lively, with fans cheering and banners visible in the background. The scoreboard and advertisements for sponsors like Hisense and BYJU’S are clearly visible around the field.\\[0.5em]

\includegraphics[width=\textwidth]{fig/02_0.7_0.png} \\
\vspace{1pt}
\footnotesize \textbf{TTF $t{=}0.7$ (first):} The video shows a soccer match between Qatar and Senegal, with Senegal leading 2-0 at the 77:28 mark. The camera pans across the field, capturing players in red and white uniforms as they move around the center circle. The stadium is filled with spectators, and advertisements for Hisense and BYU'S are visible along the sidelines. The game appears to be in the later stages, with both teams actively positioning themselves for the next play.\\[0.5em]

\includegraphics[width=\textwidth]{fig/02_0.7_-1.png} \\
\vspace{1pt}
\footnotesize \textbf{TTF $t{=}0.7$ (last):} The video shows a soccer match between Qatar and Senegal, with Senegal leading 2-0 at the 77:28 mark. The game is taking place in a large stadium filled with spectators. Players from both teams are positioned around the center circle, with Senegal in white and Qatar in red. The referee is visible near the center, and the scoreboard is displayed at the top of the screen. The atmosphere appears lively, with fans in the stands cheering and waving flags. The field is well-maintained, and the players are actively moving, suggesting the game is in progress.
\\[0.5em]

\includegraphics[width=\textwidth]{fig/02_0.7_r0.png} \\
\vspace{1pt}
\footnotesize \textbf{TTF $t{=}0.7$ ($1\times 1$ window):} The video shows a soccer match between Qatar and Senegal, with Senegal leading 2-0 at the 77:28 mark. The players are positioned around the center circle, with Senegal in white and Qatar in red. The stadium is filled with spectators, and the atmosphere appears lively. The camera pans slightly to follow the movement of the ball as players from both teams prepare for the next play.\\[0.5em]

\includegraphics[width=\textwidth]{fig/02_0.8.png} \\
\vspace{1pt}
\footnotesize \textbf{TTF $t{=}0.8$ (auto.):} The video shows a soccer match between Qatar and Senegal, with Senegal leading 2-0 at the 77:28 mark. The game is taking place in a large stadium filled with spectators. Players from both teams are positioned across the field, with Qatar in red and Senegal in white. The ball is in play near the center circle, and players are actively moving around, preparing for the next phase of play. The atmosphere is lively, with fans cheering and banners visible in the stands. The scoreboard and advertisements for sponsors like Hisense and BYU'S are clearly visible around the field.\\[0.5em]

\bottomrule
\end{tabular}
\caption{Qualitative comparison of video descriptions under different TTF configurations against the baseline using Qwen3-VL-8B-Instruct.} 
\end{table*}

\begin{table*}[!ht] 
\centering  
\small
\begin{tabular}{p{\textwidth}}  
\toprule

\includegraphics[width=\textwidth]{fig/04_baseline.png} \\
\vspace{1pt}
\footnotesize \textbf{Baseline:} In this League of Legends gameplay clip, a player controlling a character (likely Katarina or a similar assassin) is farming minions near their turret in the bottom lane. The character is using abilities to attack minions and gain gold and experience. The player's health and mana are shown at the bottom of the screen, along with their inventory and abilities. The minimap in the bottom right corner shows the positions of allied and enemy players. The game is in the late game phase, as indicated by the high kill count and the player's level. The player is focused on farming and building up their resources before engaging in team fights or pushing the lane.\\[0.5em] 

\includegraphics[width=\textwidth]{fig/04_0.7_adaptive.png} \\
\vspace{1pt}
\footnotesize \textbf{TTF $t{=}0.7$ (auto.):} The video shows a League of Legends gameplay scene where a player, controlling a character near the enemy base, is engaged in a battle. The character is attacking enemy minions and structures while managing their abilities and health. The player’s HUD displays stats like gold, experience, and item inventory, indicating an active and strategic gameplay moment. The environment is the top lane near the enemy turret, with visual effects from abilities and minions adding to the dynamic action.\\[0.5em]

\includegraphics[width=\textwidth]{fig/04_0.7_0.png} \\
\vspace{1pt}
\footnotesize \textbf{TTF $t{=}0.7$ (first):} In this League of Legends gameplay clip, a player controlling a champion (likely Katarina or a similar assassin) is in the bottom lane, engaging with enemy minions and a turret. The player is at level 15 and has a high kill-to-death ratio (21/5/4), indicating strong performance. The champion is using abilities to attack minions and possibly harass the enemy turret, while the game’s HUD shows stats like gold, experience, and cooldowns. The player’s focus seems to be on farming and maintaining pressure in the lane.\\[0.5em]

\includegraphics[width=\textwidth]{fig/04_0.7_-1.png} \\
\vspace{1pt}
\footnotesize \textbf{TTF $t{=}0.7$ (last):} In this League of Legends gameplay clip, a player controlling a champion (likely Katarina or a similar assassin) is positioned near the enemy turret in the bottom lane. The player is actively farming minions and managing their abilities, as indicated by the “cannot cast that right now!” message, likely due to cooldowns. The HUD shows the player has 15 kills, 5 deaths, and 4 assists, with a high level (15) and substantial gold. The player is focused on securing last-hitting minions and maintaining pressure near the turret, while the minimap indicates they are in the bottom lane with allies nearby. The scene is set in the early to mid-game phase, with the player aiming to control the lane and push toward the enemy base.\\[0.5em]

\includegraphics[width=\textwidth]{fig/04_0.7_r0.png} \\
\vspace{1pt}
\footnotesize \textbf{TTF $t{=}0.7$ ($1\times 1$ window):} In this League of Legends gameplay clip, a player controlling a champion is positioned near their turret in the bottom lane, engaging with enemy minions and possibly preparing for a team fight or objective. The HUD shows the player’s stats, including level, gold, and abilities, while the minimap indicates nearby allies and enemies. The environment is the typical Summoner’s Rift map, with lush greenery and stone pathways. The player appears to be in a defensive or supportive role, managing their resources and positioning strategically.
\\[0.5em]

\includegraphics[width=\textwidth]{fig/04_0.8.png} \\
\vspace{1pt}
\footnotesize \textbf{TTF $t{=}0.8$ (auto.):} In this League of Legends gameplay clip, a player controlling a champion is positioned near their turret in the bottom lane. The champion is actively farming minions and managing their abilities, as indicated by the on-screen text “cannot cast that right now!” which suggests they are on cooldown. The player’s health, mana, and item inventory are visible, showing they are in the mid-to-late game phase. The environment is the typical Summoner’s Rift map, with turrets, minions, and terrain features like trees and bushes. The player appears to be focusing on sustaining their position and preparing for future engagements.
\\[0.5em]

\bottomrule
\end{tabular}
\caption{Qualitative comparison of video descriptions under different TTF configurations against the baseline using Qwen3-VL-8B-Instruct.} 
\end{table*}

\begin{table*}[!ht] 
\centering  
\small
\begin{tabular}{p{\textwidth}}  
\toprule

\includegraphics[width=\textwidth]{fig/01_baseline.png} \\
\vspace{1pt}
\footnotesize \textbf{Baseline:} A person in a green dinosaur costume walks across a busy street at night, holding a small dog. Cars pass by in the background, and the bright lights of a city skyline are visible.
 \\[0.5em] 

\includegraphics[width=\textwidth]{fig/01_0.7_adaptive.png} \\
\vspace{1pt}
\footnotesize \textbf{TTF $t{=}0.7$ (auto.):} A person dressed in a green dinosaur costume walks across a busy street at night in Las Vegas, holding a small dog. Cars pass by in the background, and the bright lights of the city illuminate the scene.
 \\[0.5em]

\includegraphics[width=\textwidth]{fig/01_0.7_0.png} \\
\vspace{1pt}
\footnotesize \textbf{TTF $t{=}0.7$ (first):} A person dressed in a green dinosaur costume walks across a busy street at night in Las Vegas, holding a small dog. Cars pass by in the background, and the bright lights of the city illuminate the scene. \\[0.5em]

\includegraphics[width=\textwidth]{fig/01_0.7_-1.png} \\
\vspace{1pt}
\footnotesize \textbf{TTF $t{=}0.7$ (last):} A person in a green dinosaur costume walks across a busy street at night in Las Vegas, holding a small dog dressed in a matching costume. Cars pass by in the background, with bright lights from the surrounding buildings illuminating the scene.\\[0.5em]

\includegraphics[width=\textwidth]{fig/01_0.7_r0.png} \\
\vspace{1pt}
\footnotesize \textbf{TTF $t{=}0.7$ ($1\times 1$ window):} A person in a green dinosaur costume walks across a busy street at night in Las Vegas, holding a small dog. Cars pass by in the background, with the bright lights of the city illuminating the scene. \\[0.5em]

\includegraphics[width=\textwidth]{fig/01_0.8.png} \\
\vspace{1pt}
\footnotesize \textbf{TTF $t{=}0.8$ (auto.):} A person in a green dinosaur costume walks across a busy street at night in Las Vegas, holding a small dog. Cars pass by in the background, with the bright lights of the city illuminating the scene.
\\[0.5em]

\bottomrule
\end{tabular}
\caption{Qualitative comparison of video descriptions under different TTF configurations against the baseline using Qwen3-VL-8B-Instruct.} 
\end{table*}

\begin{table*}[!ht] 
\centering  
\small
\begin{tabular}{p{\textwidth}}  
\toprule

\includegraphics[width=\textwidth]{fig/03_baseline.png} \\
\vspace{1pt}
\footnotesize \textbf{Baseline:} A man in a red shirt is boarding a police helicopter. He is carrying a bag and is walking towards the open door of the helicopter. The helicopter is parked on a helipad and the sun is shining brightly. The man is silhouetted against the bright sky.\\[0.5em] 

\includegraphics[width=\textwidth]{fig/03_0.7_adaptive.png} \\
\vspace{1pt}
\footnotesize \textbf{TTF $t{=}0.7$ (auto.):} A man in a red shirt is boarding a dark blue helicopter marked "Bundespolizei" on a sunny day. He walks toward the open door, reaches in, and steps inside. The camera follows his movement as he enters the helicopter.\\[0.5em]

\includegraphics[width=\textwidth]{fig/03_0.7_0.png} \\
\vspace{1pt}
\footnotesize \textbf{TTF $t{=}0.7$ (first):} A man in a red shirt approaches a dark blue Bundespolizei helicopter parked on a helipad. He reaches for the door handle and begins to climb into the helicopter. The sun is shining brightly, creating lens flares in the camera.\\[0.5em]

\includegraphics[width=\textwidth]{fig/03_0.7_-1.png} \\
\vspace{1pt}
\footnotesize \textbf{TTF $t{=}0.7$ (last):} A man in a red shirt approaches a dark blue Bundespolizei helicopter parked on a helipad. He reaches for the door handle and begins to climb into the helicopter. The sun shines brightly, casting a lens flare over the scene.\\[0.5em]

\includegraphics[width=\textwidth]{fig/03_0.7_r0.png} \\
\vspace{1pt}
\footnotesize \textbf{TTF $t{=}0.7$ ($1\times 1$ window):} A man in a red shirt approaches a dark blue Bundespolizei helicopter parked on a helipad. He reaches for the door handle and begins to step into the helicopter. The sun shines brightly, casting a lens flare across the scene. \\[0.5em]

\includegraphics[width=\textwidth]{fig/03_0.8.png} \\
\vspace{1pt}
\footnotesize \textbf{TTF $t{=}0.8$ (auto.):} A man in a red shirt approaches a dark blue Bundespolizei helicopter parked on a helipad. He opens the side door and steps inside, then closes the door behind him. The sun shines brightly, creating lens flares.\\[0.5em]

\bottomrule
\end{tabular}
\caption{Qualitative comparison of video descriptions under different TTF configurations against the baseline using Qwen3-VL-8B-Instruct.} 
\end{table*}

\end{document}